\documentclass[journal]{IEEEtran}
\usepackage{amsmath,amsfonts}
\usepackage{array}
\usepackage{balance}
\usepackage{multirow}
\usepackage{textcomp}
\usepackage{stfloats}
\usepackage{url}
\usepackage{verbatim}
\usepackage{graphicx}
\usepackage{cite}
\usepackage[font=small]{caption}
\usepackage[font=small]{subcaption}

\usepackage{bm}
\usepackage{bbm}
\usepackage{xcolor}
\definecolor{grey}{rgb}{0.5,0.5,0.5}
\usepackage[normalem]{ulem}
\useunder{\uline}{\ul}{}
\usepackage[ruled,linesnumbered]{algorithm2e}

\begin{document}

\title{Diffusion-Modeled Reinforcement Learning for Carbon and Risk-Aware Microgrid Optimization}

\author{Yunyi Zhao,~\IEEEmembership{Student Member,~IEEE},~Wei Zhang,~\IEEEmembership{Member,~IEEE,}~Cheng Xiang,~\IEEEmembership{Member,~IEEE,}\\\quad\quad Hongyang Du,~\IEEEmembership{Member,~IEEE,}~Dusit Niyato,~\IEEEmembership{Fellow,~IEEE,} and~Shuhua Gao,~\IEEEmembership{Member,~IEEE}
\thanks{Manuscript received January 1, 2025; revised January 1, 2025; accepted January 1, 2025. This research was supported in part by MTC Individual Research Grants (IRG) (Award M23M6c0113), A*STAR under its MTC Programmatic (Award M23L9b0052), and the National Research Foundation Singapore and DSO National Laboratories under the AI Singapore Programme (AISG Award No: AISG2-GC-2023-006). (\textit{Corresponding author: Wei Zhang})}
\thanks{Yunyi Zhao is with both the Information and Communications Technology Cluster, Singapore Institute of Technology, Singapore 138683, and the Department of Electrical and Computer Engineering, National University of Singapore, Singapore 119077 (e-mail: yunyi.zhao@singaporetech.edu.sg).}
\thanks{Wei Zhang is with the Information and Communications Technology Cluster, Singapore Institute of Technology, Singapore 138683 (e-mail: wei.zhang@singaporetech.edu.sg).}
\thanks{Cheng Xiang is with the Department of Electrical and Computer Engineering, National University of Singapore, Singapore 119077 (e-mail: elexc@nus.edu.sg).}
\thanks{Hongyang Du is with the Department of Electrical and Electronic Engineering, University of Hong Kong, Hong Kong SAR, China (e-mail: duhy@eee.hku.hk).}
\thanks{Dusit Niyato is with the College of Computing and Data Science, Nanyang Technological University, Singapore 639798 (e-mail: dniyato@ntu.edu.sg).}
\thanks{Shuhua Gao is with the School of Control Science and Engineering, Shandong University, Jinan, China (e-mail: shuhuagao@sdu.edu.cn).}
}



\maketitle


\begin{abstract}
This paper introduces \textsc{DiffCarl}, a \underline{diff}usion-modeled \underline{ca}rbon- and \underline{r}isk-aware reinforcement \underline{l}earning algorithm for intelligent operation of multi-microgrid systems. With the growing integration of renewables and increasing system complexity, microgrid communities face significant challenges in real-time energy scheduling and optimization under uncertainty. \textsc{DiffCarl} integrates a diffusion model into a deep reinforcement learning (DRL) framework to enable adaptive energy scheduling under uncertainty and explicitly account for carbon emissions and operational risk. By learning action distributions through a denoising generation process, \textsc{DiffCarl} enhances DRL policy expressiveness and enables carbon- and risk-aware scheduling in dynamic and uncertain microgrid environments. Extensive experimental studies demonstrate that it outperforms classic algorithms and state-of-the-art DRL solutions, with 2.3$-$30.1\% lower operational cost. It also achieves 28.7\% lower carbon emissions than those of its carbon-unaware variant and reduces performance variability. These results highlight \textsc{DiffCarl} as a practical and forward-looking solution. Its flexible design allows efficient adaptation to different system configurations and objectives to support real-world deployment in evolving energy systems.
\end{abstract}

\begin{IEEEkeywords}
Diffusion models, generative articial intelligence, multi-microgrids, carbon aware, energy scheduling, risk-sensitive, deep reinforcement learning
\end{IEEEkeywords}

\section{Introduction}
\IEEEPARstart{I}{n} recent years, microgrids have gained increasing attention in many nations with increasing utilization of renewable energy sources as substitutes of traditional fossil energy. Such transition not only brings economic benefits but also contributes to environmental sustainability by reducing emissions of carbon dioxide and other greenhouse gases. Microgrids adopt a localized and decentralized approach to manage energy with higher resilience and flexibility than traditional centralized grids. In addition to renewable distributed generation (RDG), microgrids often integrate energy storage systems (ESS), controllable diesel generation (CDG), and loads. While conventional microgrids typically involve control within a bounded system, recent developments have expanded this concept. The notion of a microgrid community (MGC) \cite{article2016} has been introduced to represent decentralized and cooperative energy systems that span multiple microgrids for achieving collective energy scheduling and optimization goals.

At the system level, the goal of operating microgrids and MGC is to maximize efficiency, reduce costs, and minimize environmental impact while ensuring stability and reliability. Practical energy scheduling faces key challenges, including the uncertainty of renewable generation, demand variability, and coordination of heterogeneous components under constraints like limited communication and decentralized control. Operational limits, such as ramping rates, storage state-of-charge, and network constraints, add non-linearity and complicate real-time optimization \cite{LI2018974}. In large-scale MGC, the growing number of decision variables and amplified uncertainty further increase complexity, making effective inter-microgrid coordination essential. These challenges demand robust, scalable, and adaptive solutions responsive to dynamic conditions. 

Numerous studies have addressed energy scheduling and optimization in microgrids. Early approaches formulated the problem as mixed-integer nonlinear programming (MINLP) \cite{alipour2017minlp} or as mixed-integer linear programming (MILP) with linearization for tractability \cite{hosseinzadeh2015robust, thomas2018optimal}. To incorporate real-time information, model predictive control (MPC) was introduced as an extension of MILP-based frameworks \cite{parisio2014model, petrollese2016real}, offering some robustness but still limited to single or centralized microgrids. To improve scalability, hierarchical and distributed frameworks have been proposed, such as a two-level hierarchical MILP for negotiation among agents \cite{tian2015hierarchical} and a distributed MILP for coordinating interconnected microgrids \cite{liu2022distributed}. Despite these advances, numerical optimization remains computationally expensive and relies on accurate predictions of renewables, loads, and markets, which is a significant challenge in practice.

Recently, data-driven machine learning (ML), particularly deep reinforcement learning (DRL), has been explored to improve optimization performance and efficiency. The problem is often cast as a Markov decision process (MDP), where agents learn near-optimal policies by interacting with the environment, avoiding the need for precise forecasts. DRL has shown fast convergence and strong performance in applications like battery operation optimization, demand response, and energy management \cite{bui2019double, amer2022drl}. However, early value-based methods like Q-learning and DQN struggle with the large state and action spaces typical in microgrids. To address this, imitation learning (IL) has been used to leverage expert demonstrations for faster training and near-optimality \cite{gao2021online, zhao2024online, gao2025multi}, but suffers from overfitting and distributional shift due to reliance on expert policy quality \cite{kostrikov2019imitation}. More advanced methods such as DDPG \cite{dolatabadi2022novel}, SAC \cite{ren2024data}, PPO \cite{shi2023coordinated}, and TD3 \cite{zhang2024scalable} have since been developed.

Despite these advances, energy scheduling in microgrid clusters (MGC) remains challenging due to dynamic environments, combinatorial action spaces, and multi-agent coordination. Recently, generative AI (GenAI) has emerged as a promising approach, enhancing robustness, interpretability, and adaptability under uncertainty. GenAI shifts from deterministic to probabilistic and adaptive decision-making, augmenting DRL by modeling distributions over actions rather than fixed policies. Its ability to emulate experts while generalizing to novel conditions makes it a valuable tool for intelligent energy scheduling. In \cite{du_diffusion-based_2023}, GenAI integrated with SAC improved performance in a non-energy domain. In \cite{zhang_two-step_2024}, a GenAI-based DRL algorithm for multi-energy scheduling was proposed but still derived a deterministic policy, limiting robustness under high uncertainty. These limitations underscore the need to further improve GenAI-based approaches for reliable scheduling in complex energy systems.


In this paper, we leverage the strength of GenAI and improve DRL for MGC energy scheduling and optimization. By formulating decision-making as a denoising generation process, diffusion policies offer both diversity and constraint-awareness, making them well-suited for dynamic, uncertain, and real-time energy environments. We propose \textsc{DiffCarl}: \underline{diff}usion-modeled \underline{ca}rbon and \underline{r}isk-Aware Reinforcement \underline{l}earning, a novel framework that addresses key limitations in existing solutions in two main aspects. First, \textsc{DiffCarl} leverages diffusion modeling to learn expressive and flexible policies capable of handling complex MGC dynamics and outperforming conventional DRL algorithms. Second, unlike many traditional solutions focusing solely on minimizing energy cost, \textsc{DiffCarl} explicitly incorporates carbon intensity and operational risk into its decision-making. This integrated design is driven by the practical need for sustainable and resilient microgrids, where operations shall align with decarbonization goals and safety-critical constraints. Overall, \textsc{DiffCarl} is a practical and forward-looking solution for energy scheduling under increasing system complexity and uncertainty. The main contributions of this paper are summarized as follows.

\begin{enumerate}
    \item We formulate an energy scheduling and optimization problem and an interactive learning environment for MGC, explicitly incorporating carbon emission and operational risks into the formulation.
    \item We propose \textsc{DiffCarl} algorithm, which integrates diffusion model in DRL to optimize MGC energy schedules with improved performance in spite of the uncertainty of renewable, loads, and the market.
    \item We apply carbon- and risk-awareness to \textsc{DiffCarl} algorithm, which effectively minimizes carbon emission and reduces the risk of high operational cost, thereby satisfying the user's need of balancing cost efficiency, sustainability and manageable risk exposure.
    \item We conduct extensive experiments to evaluate the performance of \textsc{DiffCarl}. The results show that \textsc{DiffCarl} achieves lower operational cost, less carbon emission and better risk management compared with several classic and state-of-the-art DRL algorithms.
\end{enumerate}

The remainder of this paper is organized as follows. Section \ref{sec-diffcarl} details the system architecture and the proposed algorithm \textsc{DiffCarl}. Section \ref{sec-exp} reports the simulation studies and provides a comprehensive discussion. Finally, Section \ref{sec-conclusion} concludes the paper.

\section{Methodology}
\label{sec-diffcarl}
\begin{figure}[!t]
    \centering
    \includegraphics[width=0.95\linewidth]{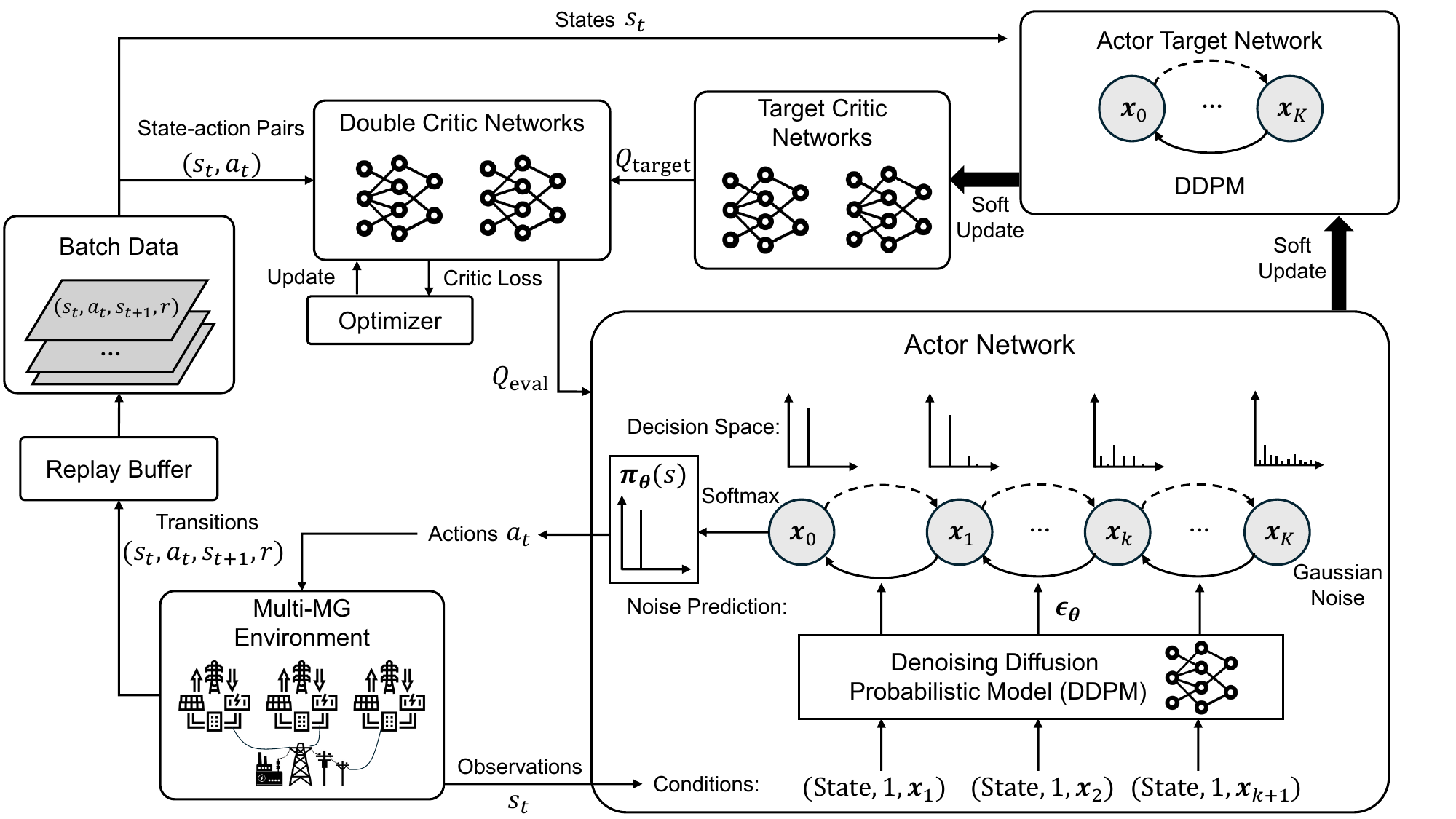}
    \caption{The framework of \textsc{DiffCarl} with diffusion-modeled actor network.}
    \label{fig:systemarch}
\end{figure}

In this paper, we consider a classic microgrid cost minimization problem \cite{parisio2014model, gao2021online, tian2015hierarchical} and incorporate carbon-awareness into a MILP-based optimization problem. Typically, a classic microgrid model consists of controllable distributed generators (CDGs), non-controllable renewable distributed generators (RDGs), and energy storage systems (ESSs); generally, the microgrids' power is exchangeable with the utility grid (UG) in the grid-connected mode and nonexchangeable in the islanded mode. Afterwards, we work on combining generative models in reinforcement learning (RL). Traditional RL is inefficient in real-world applications due to its slow convergence and is vulnerable to overestimation and risks in real world applications. In this section, we aim to solve the formulated optimization problem for MGC energy management and we present our proposed algorithm \textsc{DiffCarl}. \textsc{DiffCarl} is a diffusion model-based DRL algorithm which follows the well-known SAC algorithm \cite{haarnoja2018soft}, where the actor network is replaced by a diffusion-modeled action generation module to enhance the actor's exploration capability. The system framework is described in Fig. \ref{fig:systemarch} and we present the technical details below.

\subsection{The Basics of RL}

In the formulated problem, the microgrid community will be operated in a way of sequential decision-making process. The MDP is represented by a 4-tuple $(\mathcal{S}, \mathcal{A}, \mathcal{P}, \mathcal{R})$, where $\mathcal{S}$ stands for a finite set of observable states of an environment; $\mathcal{A}$ is a finite set of actions; $\mathcal{P}$ is the transition probability function and we have $\sum_{s'\in \mathcal{S}}\mathcal{P}(s_t, a_t, s')=1$, where $s_t$ is the current state, $a_t$ is the action taken for time $t$, and $s'$ represents a possible state of $s_{t+1}$; $\mathcal{R}$ is the reward function; and $\gamma$ is the discount factor. A RL agent learns a policy $\pi(a_t|s_t)$ with optimized decision making to maximize the cumulative reward. At each time $t$, the agent generates and executes action $a_t$ according to the policy $\pi$ based on state $s_t$, and receives a reward $r_t$. The agent keeps observing new states, making actions, and measuring reward, and gradually optimizes the policy $\pi$ based on continuous interactions with the environment. The details of the tuple are as follows.

\subsubsection{State}
The state of the MGC at time $t$ includes crucial microgrid information including load, renewable energy generation, electricity prices, and ESS SoC, constructed as
\begin{equation}\label{eq3-1:rl-state}
    s_t = \big[ P_{\textrm{LD}}(t), P_{\textrm{RDG}}(t), \rho(t), \text{SoC}(t)\big].
\end{equation}

\subsubsection{Action}
The power generation or consumption of several major MGC components, including ESS, CDG and load shedding power (LS), is usually controllable. Therefore, they become the focus of the action vector in decision making and control, as
\begin{equation}\label{eq3-2:rl-action}
    a_t = \big[ P_{\textrm{ESS}}(t), P_{\textrm{CDG}}(t), P_{\textrm{LS}}(t) \big].
\end{equation}

\subsubsection{Transition Probability} 
The transition probability from state $s$ at time $t-1$ to state $s'$ at time $t$ is denoted as $\mathcal{P}_{s\rightarrow s^{\prime}} = P\left(s_{t+1}|s_{t}, a_{t}\right)$, given that action $a_t$ is taken at time $t$. In this paper, we redesign the SAC framework of RL by utilizing the generative capability of diffusion models and developing a diffusion-based action generator based on the latest state. The details will be discussed in the following subsections.

\subsubsection{Risk-Aware Reward} 
Risk shall be accounted for in many real-world decision-making applications to avoid potentially significant losses or unsafe system behaviors. Often, the importance of a low-risk system even outweighs certain sacrifices of expected returns, as several risk-aware RL algorithms have been proposed in \cite{shen2014risk, enders2024risk}. Specifically for our MGC energy management, risk-awareness is highly important for grid operators to avoid unacceptable spike cost and carbon mission. As such, we incorporate the risk factor into our solution, by optimizing the policy as
\begin{equation}\label{eq3-3-1:rl-risk}
    \pi^{*} = \arg\max_{\pi} \left\{\mathbb{E}_{\pi}\left[ \sum_{t=1}^{\infty} \gamma r(s_t,a_t)\right] - \lambda_{\textrm{risk}} R(\pi)\right\},
\end{equation}
where we maintain a typical RL reward term with discount factor $\gamma$ in the first term and further introduce the second term $R(\pi)$ as the risk measurement function associated with policy $\pi$. The two terms are balanced by a control parameter $\lambda_{\textrm{risk}}$, where a positive value indicates a risk-averse strategy and vice verse. Conditional Value at Risk (CVaR) is commonly used \cite{eriksson2019epistemic,ma2025dsac} as the risk term $R(\pi)$ to penalize extreme costs, thereby focusing on eliminating the worst-case cost scenarios

\begin{equation}\label{eq3-3-2:cvar-term}
    R(\pi) = \mathbb{E}_{\pi}\left[\textrm{Cost} | \textrm{Cost} > \textrm{VaR}_{\alpha_{\textrm{CVaR}}}(\textrm{Cost})\right],
\end{equation}

\noindent where $\alpha_{\textrm{CVaR}}$ stands for the percentage of worst-case scenarios to evaluate. For example if $\alpha_{\textrm{CVaR}} = 0.95$, Eq. \eqref{eq3-3-2:cvar-term} measures the $5\%$ CVaR to evaluate the average cost in bad days when the cost is higher than $95\%$ of the others during. Such risk term helps to optimize the worst-case robustness and promote safety and reliability.

\subsection{The Basics of Diffusion Models}
The actor network of \textsc{DiffCarl} is designed based on a diffusion model. Diffusion models are originally designed for image generation. It involves a forward process where noise is gradually added to the data until data becomes Gaussian noise, from which a reverse process starts to generate data that aligns with the distribution of the original data. Inspired by its inherent generative ability in denoising process, we aim to design a diffusion-based optimizer to enhance precision in decision making problems based on environment conditions. The diffusion-based actor network consistently evaluates and optimizes current policy by interacting with the MGC environment. In the reverse process, the diffusion-based actor $\pi_{\bm{\theta}}(\cdot)$ generates the optimal decision based on conditioning information, which can be viewed as a denoising process starting from Gaussian noise to gradually recover the optimal decision distributions. This design improves the adaptability of DRL and leverages the generative strengths of diffusion models, underscoring their complementary roles in decision-making. Therefore, such combination results in an effective diffusion-based policy optimizer which outperforms conventional DRL approaches in complex and dynamic combinatorial optimization problems \cite{du_diffusion-based_2023, zhang_two-step_2024}. We present technical details of the diffusion model in this part.

\subsubsection{Forward Process} 
Let $\bm{x}_0$ be the desired action generated by the actor network. $\bm{x}_0$ is a target data sample, which is sampled by a probability distribution of decisions under a condition and the process can be represented as
\begin{equation}\label{eq3-4:diff-forward-x0}
    \bm{x}_0 = \pi_{\bm{\theta}}(s) \sim \mathbb{R}^{\mathcal{A}},
\end{equation}
\noindent where $s$ is the observed state, and $\bm{\theta}$ is the model parameters of the diffusion model. To simply the presentation, we first assume that $\bm{x}_0$ is known. The forward process is performed to add Gaussian noise to the original distribution step by step, deriving a sequence of noised data samples $\bm{x}_1, \bm{x}_2, \ldots, \bm{x}_K$, where $K$ is the total diffusion steps. We represent the transition from $\bm{x}_{k-1}$ to $\bm{x}_{k}$ as $q(\bm{x}_k|\bm{x}_{k-1})$, which can be written as
\begin{equation}
\label{eq3-5:diff-forward-q}
    q(\bm{x}_k|\bm{x}_{k-1})=\mathcal{N}(\bm{x}_k|\sqrt{1-\beta_k}\bm{x}_{k-1}, \beta_k \bm{I}),
\end{equation}
where $\beta_k$ is the forward process variance for $1\leq k \leq K$, and $\bm{I}$ is an identity matrix. We refer to \cite{ho2020denoising} and follow the variational posterior scheduler to schedule $\beta_t$ at each step as
\begin{equation}
\label{eq3-6:diff-forward-beta}
    \beta_k = 1 - e^{-\frac{\beta_{\min}}{K} - \frac{2k - 1}{2K^2}(\beta_{\max} - \beta_{\min})},
\end{equation}
where $\beta_{\min}$ and $\beta_{\max}$ are hyper-parameters. The forward process is essentially a Markov process where $\bm{x}_{k}$ is only dependent on $\bm{x}_{k-1}$, and the process creates a mathematical relation between $\bm{x}_0$ and $\bm{x}_k$ as
\begin{equation}\label{eq3-7:diff-forward-xt}
    \bm{x}_k = \sqrt{\bar{\alpha}_k} \bm{x}_0 + \sqrt{1-\bar{\alpha}_k} \bm{\epsilon},
\end{equation}

\noindent where we define $\alpha_k = 1-\beta_k$, $\bar{\alpha}_k = \prod_{i=1}^{k} \alpha_i$, and $\bm{\epsilon} \sim \mathcal{N}(0,\bm{I})$ as a standard Gaussian noise. The implication is that, with the increase of diffusion step $k$, $\bm{x}_k$ gradually becomes a sample defined by a standard normal distribution at step $K$.

All above derivations are based on the assumption that $\bm{x}_0$ is known, which, however, may not be feasible for many real-world problems, where the distribution of optimal actions is not fully clear or available. Indeed, the forward process cannot directly obtain the distribution of optimal actions with respect to the observed state. However, it builds up the connection between original and noised data, and the connection is vital for the analysis of the reverse process. For example, we can obtain the reverse conditional probability $q(\bm{x}_{k-1}|\bm{x}_k, \bm{x}_0)$ conditioned on $\bm{x}_0$ as
\begin{equation}
\label{eq3-8:diff-forward-reverse-condition}
    q(\bm{x}_{k-1}|\bm{x}_k, \bm{x}_0) = \mathcal{N}\left(\bm{x}_{k-1}; \tilde{\bm{\mu}}(\bm{x}_k, k), \tilde{\beta}_k \bm{I}\right),
\end{equation}
where $\tilde{\bm{\mu}}(\bm{x}_k, k)$ and $\tilde{\beta}_k$ represent the mean and variance of the conditional posterior $q(\bm{x}_{k-1}|\bm{x}_k, \bm{x}_0)$, respectively. They can be calculated from Gaussian conditioning rules as
\begin{subequations}
\label{eq3-10:diff-forward-reverse-param}
    \begin{align}
        \tilde{\bm{\mu}}(\bm{x}_k, k) & =\frac{\sqrt{\alpha_k}(1 - \bar{\alpha}_{k-1})}{1 - \bar{\alpha}_k} \bm{x}_k + \frac{\sqrt{\bar{\alpha}_{k-1}} \beta_k}{1 - \bar{\alpha}_k} \bm{x}_0, \\
        \tilde{\beta}_k & = \frac{1 - \bar{\alpha}_{k-1}}{1 - \bar{\alpha}_k} \cdot \beta_k.
    \end{align}
\end{subequations}

\subsubsection{Reverse Process} 
The objective of the reverse process is to denoise a pure noise sample $\bm{x}_K \sim \mathcal{N}(0,\bm{I})$ to acquire the target sample $\bm{x}_0$, by removing the noise step by step. Let $p(\bm{x}_{k-1}|\bm{x}_k)$ be the transition of noise reduction from step $k$ to step $k-1$. As the probability transition function cannot be derived directly, we propose to use a model $p_{\bm{\theta}}$ to approximate the function, and we have
\begin{equation}
\label{eq3-11:diff-reverse-model-ptheta}
    p_{\bm{\theta}}(\bm{x}_{k-1}|\bm{x}_k) = \mathcal{N}\left(\bm{x}_{k-1}; \bm{\mu}_{\bm{\theta}}(\bm{x}_k, k, s), \Sigma_{\bm{\theta}}(\bm{x}_k, k, s)\right),
\end{equation}
\noindent where $s$ is the observed state. Our target is to train a model $p_{\bm{\theta}}$ properly, so that the variables in Eq. \eqref{eq3-10:diff-forward-reverse-param} can be well approximated with $\bm{\mu}_{\bm{\theta}}(\bm{x}_k, k, s) \rightarrow \tilde{\bm{\mu}}(\bm{x}_k, k, s)$ and $\Sigma_{\bm{\theta}}(\bm{x}_k, k, s) \rightarrow \tilde{\beta}_k$. Based on Eqs. \eqref{eq3-6:diff-forward-beta} and \eqref{eq3-10:diff-forward-reverse-param}, $\Sigma_{\bm{\theta}}(\bm{x}_k, k, s)= \tilde{\beta}_k \bm{I}$ is a series of deterministic constant values controlled by variational posterior scheduler. Furthermore, according to Eq. \eqref{eq3-7:diff-forward-xt}, the reconstruction of $\bm{x}_0$ can be obtained by merging multiple Gaussian signals using reparameterization as
\begin{equation}\label{eq3-13:diff-reverse-model-x0}
    \bm{x}_0  = \frac{1}{\sqrt{\bar{\alpha}_k}} \bm{x}_k + \sqrt{\frac{1}{\bar{\alpha}_k} - 1} \cdot \tanh \left(\bm{\epsilon}_{\bm{\theta}}(\bm{x}_k, k, s)\right),
\end{equation}
\noindent where $\bm{\epsilon}_{\bm{\theta}}(\bm{x}_k, k, s)$ is a multi-layer neural network model parameterized by $\bm{\theta}$ which approximates the noises conditioned on state $s$ to be denoised. We use the derived $\bm{x}_0$ to calculate Eq. \eqref{eq3-10:diff-forward-reverse-param} and we can estimate the mean $\tilde{\bm{\mu}}(\bm{x}_k, k, s)$ by
\begin{equation}\label{eq3-14:diff-reverse-model-mu-new}
    \tilde{\bm{\mu}}(\bm{x}_k, k, s)  = \frac{1}{\sqrt{\alpha_k}}\left( \bm{x}_k - \frac{\beta_k \cdot \tanh \bm{\epsilon}_{\bm{\theta}}(\bm{x}_k, k, s)}{\sqrt{1-\bar{\alpha}_k}} \right).
\end{equation}

Based on the noise approximation model $\bm{\epsilon}_{\bm{\theta}}(\bm{x}_k, k, s)$, we sample $\bm{x}_{k-1}$ from Eqs. \eqref{eq3-11:diff-reverse-model-ptheta} and \eqref{eq3-14:diff-reverse-model-mu-new}. The properties of a Markov process allows us to derive the generation distribution $p_{\bm{\theta}}(\bm{x}_0)$ as
\begin{equation}
\label{eq3-15:diff-reverse-model-ptheta-x0}
    p_{\bm{\theta}}(\bm{x}_0) = p(\bm{x}_{K})\prod_{k=1}^{K}p_{\bm{\theta}}(\bm{x}_{k-1}|\bm{x}_k),
\end{equation}
\noindent where $p(\bm{x}_{K})$ is a standard Gaussian distribution as the start of the reverse process. Furthermore, we reparameterize the noise term instead to predict $\bm{x}_{k-1}$ as
\begin{equation}\label{eq3-16:diff-reparameterize}
    \bm{x}_{k-1} = \mathcal{N}\big(\bm{x}_{k-1}; \bm{\mu}_{\bm{\theta}}(\bm{x}_k, k, s), \Sigma_{\bm{\theta}}(x_k, k, s)\big).
\end{equation}

\subsection{\textsc{DiffCarl} Algorithm Description}
\textsc{DiffCarl} is RL-based and the training process begins with the RL agent observing the environment to obtain initial state $s_0$. The agent will utilize the diffusion model to iteratively perform denoising and generate an action based on the current state. Specifically at each step $t$, the actor generates a probability distribution $\pi_{\bm{\theta}}(s_t)$ over all possible actions based on the observation $s_t$. Then, an action $a_t \sim \pi_{\bm{\theta}}(s_t)$ is sampled and fed into the environment. The environment processes the action and sends back a new state $s_{t+1}$ and a reward $r_t$ as the feedback to the agent. The agent records the transitions by interacting with the environment and the transitions from different time steps are saved into the replay buffer $\mathcal{B}$. 

\subsubsection{Action Entropy Regularized Policy Improvement}
One of key objectives of RL is to improve the actor network, or policy, so that optimal actions can be generated based on the observed states. Typically, the policy is gradually optimized by sampling batches of transitions from replay buffer $\mathcal{B}$ and maximizing the expectation of Q values over all actions as
\begin{equation}
\label{eq3-17:policy-improvement}
    \max_{\bm{\theta}} ~\pi_{\bm{\theta}}^{T}(s) Q^{\textrm{risk}}_{\bm{\phi}}(s) + \alpha_{\textrm{ent}} H\left(\pi_{\bm{\theta}}(s)\right),
\end{equation}
\noindent where the first term, $\pi_{\bm{\theta}}^{T}(s)$ denotes current policy and $Q^{\textrm{risk}}_{\bm{\phi}}(s)$ is a risk-adjusted Q-value function which outputs a Q-value $\bm{q}$. The second term models the entropy of the action probability $\pi_{\bm{\theta}}(s)$, and can be calculated as $H\left(\pi_{\bm{\theta}}(s)\right) = - \pi^{T}_{\bm{\theta}}(s) \log \pi_{\bm{\theta}}(s)$. $\alpha_{\textrm{ent}}$ is the temperature co-efficiency for controlling the strength of the entropy to encourage exploration.

By maximizing Eq. \eqref{eq3-17:policy-improvement}, the agent optimizes the actor network in a direction of selecting the actions with high Q-values. The parameters of the network are updated based on the gradient descent as
\begin{equation}
\label{eq3-18:actor-param-update}
    \bm{\theta}_{e+1} \leftarrow \bm{\theta}_{e} - \eta_{a} \cdot \mathbb{E}_{\left(s,a,s^{\prime},r\right) \sim \mathcal{B}} \left(\nabla_{\bm{\theta}_{e}}\left[
    \begin{aligned}
    - \alpha_{\textrm{ent}} H&\left(\pi_{\bm{\theta}}(s)\right)\\
    - \pi^{T}_{\bm{\theta}_{e}}(s) &Q^{\textrm{risk}}_{\bm{\phi}_{e}}(s) 
    \end{aligned}
    \right]\right), 
\end{equation}

\noindent where $\eta_a$ represents the learning rate of the actor network and $e$ identifies an epoch.

\subsubsection{Risk-sensitive Q-function Improvement}
Q-function network $Q_{\bm{\phi}}(s)$ as the critic is crucial for \textsc{DiffCarl} for evaluating policy $\pi_{\bm{\theta}}(s)$ and we design the risk-adjusted Q-function $Q^{\textrm{risk}}_{\bm{\phi}}(s)$ based on CVaR. In every epoch $e$, we update the Q-function effectively and utilize the temporal difference error to minimize the loss function
\begin{equation}
\label{eq3-19-1:q-function-loss}
    \mathcal{L}^{\textrm{risk}}_{Q}(\phi_{e}) = \mathbb{E}_{\left(s,a,s^{\prime},r\right) \sim \mathcal{B}}\left[\bigg(Q^{\textrm{risk}}_{\phi_{e}}(s,a) - y_{\textrm{CVaR}}\bigg)^2\right],
\end{equation}

\noindent where $Q^{\textrm{risk}}_{\bm{\phi}_{e}}(s,a)$ represents the risk-sensitive Q-value given by the critic networks corresponding to the state and action at each epoch $e$, and the CVaR-based target $y_{\textrm{CVaR}}$ is updated by

\begin{equation}
\label{eq3-19-2:q-function-update}
    y_{\textrm{CVaR}} = r + \gamma \cdot \textrm{CVaR}_{\alpha_{\textrm{CVaR}}} \left( \min_{i=1,2} \hat{Q}_{\hat{\bm{\phi}}^{i}_{e}}(s', a') - \alpha_{\textrm{ent}} \log\hat{\pi}_{\bm{\theta}_{e}}(s^{\prime})\right),
\end{equation}

\noindent where $r$ stands for the immediate reward, $\gamma$ denotes the discount factor, $i=1, 2$ represent the different critic networks, $\textrm{CVaR}_{\alpha_{\textrm{CVaR}}}(\cdot)$ is a CVaR-regulated term based on Eq. \eqref{eq3-3-2:cvar-term}, $\alpha_{\textrm{ent}} \log\hat{\pi}$ is an action entropy term, while $\hat{\pi}$ and $\hat{Q}$ stand for the target actor and critic networks with parameters $\hat{\bm{\theta}}$ and $\hat{\bm{\phi}}$, respectively.

\subsubsection{Soft-Update}
In \textsc{DiffCarl}, we adopt soft-update for the actor and critic networks to improve convergence stability during training. Instead of updating network parameters abruptly as hard-update does, parameters are updated mildly and such a smooth transition helps prevent large swings or oscillations in estimations and reduce the risk of divergence. In practice, a hyper-parameter $\tau \in (0,1]$ is used to control the update rate, and we have
\begin{subequations}
\label{eq3-20:soft-update}
    \begin{align}
    &\hat{\bm{\theta}}_{e+1} \leftarrow \tau \bm{\theta}_{e} + (1-\tau) \hat{\bm{\theta}}_{e},\\
            &\hat{\bm{\phi}}_{e+1} \leftarrow \tau \bm{\phi}_{e} + (1-\tau) \hat{\bm{\phi}}_{e},
        \end{align}
\end{subequations}

\noindent where $\hat{\bm{\theta}}$ and $\hat{\bm{\phi}}$ are updated through soft-update instead of being updated directly based on gradient descent. Overall, \textsc{DiffCarl} iteratively updates both actor and critic networks epoch by epoch until reaching convergence. We present the pseudo-code of \textsc{DiffCarl} in Algorithm \ref{algo/DiffCaRL}. The computational complexity of the proposed \textsc{DiffCarl} is $\mathcal{O}\left(E\left[C\left(V+KN_{\bm{\theta}}\right)+(d+1)\left(N_{\bm{\theta}} + N_{\phi}\right) + d\log d\right]\right)$, where $E$ stands for training episodes, $C$ is the transitions collected per episode, $V$ represents the complexity of the environment, $K$ is the denoising steps, $N_{\bm{\theta}}$ and $N_{\bm{\phi}}$ are the parameter numbers of the actor and critic networks respectively, and $d$ stands for the batch size. The space complexity of \textsc{DiffCarl} is $\mathcal{O}(2N_{\bm{\theta}} + 4N_{\phi} + B(2|\mathcal{S}| + |\mathcal{A}| + 1))$, where $B$ represents the capacity of the replay buffer $\mathcal{B}$, while $|\mathcal{S}|$ and $|\mathcal{A}|$ are the dimensions of states and actions respectively.

\begin{algorithm}[t]
    \small
    \caption{\textsc{DiffCarl}: Diffusion-Modeled Carbon and Risk-Aware Reinforcement Learning}
    \label{algo/DiffCaRL}
    \KwIn{A well-defined MGC environment $\mathcal{E}$, number of maximum training episodes $E$, Transitions per training step $C$, weight of soft update $\tau$, discount factor $\gamma$, temperature co-efficiency $\alpha$, learning rate of actor network $\eta_a$, learning rate of critic networks $\eta_c$, confidence level of CVaR $\alpha_{\textrm{CVaR}}$, batch size $d$}
    \KwOut{A \textsc{DiffCaRL} policy $\pi^*$ with parameters $\bm{\theta}^*$}

    Initialize a diffusion policy $\pi_{\bm{\theta}}$ with parameters $\bm{\theta}$, Q-function networks $Q_{\bm{\phi}}$ with parameters $\bm{\phi}$ and a replay buffer $\mathcal{B}$\;

    Initialize target network parameters $\hat{\bm{\theta}} \leftarrow \bm{\theta}$, $\hat{\bm{\phi}} \leftarrow \bm{\phi}$\;
    
    \For{ \textrm{training episode} $e \leftarrow 1$ \KwTo $E$}{

        \For{ \textrm{transition updates} $c \leftarrow 1$ \KwTo $C$}{

            Initialize a normal distribution $x_{K} \sim \mathcal{N}(\bm{0},\bm{I})$\;
            
            Observe state $s$\;
            
            \For{ \textrm{denoising step} $k \leftarrow K$ \KwTo $1$}{
              
              Scale a denoising factor $\bm{\epsilon}(\bm{x}_{k}, k, s)$ based on $x_{k}$\;
    
              Calculate the distribution $\bm{x}_{k-1}$ based on \eqref{eq3-11:diff-reverse-model-ptheta}, \eqref{eq3-14:diff-reverse-model-mu-new}, \eqref{eq3-16:diff-reparameterize}.
            }
    
            Calculate probability distribution $\pi_{\bm{\theta}}(s)$ of $\bm{x}_0$ based on a softmax function\;
    
            Select an optimal action $a$ based on $\pi_{\bm{\theta}}(s)$\;
    
            Execute $a$ and observe the environment feedback with reward $r$ and next state $s^{\prime}$\;
    
            Save the transition $(s, a, s^{\prime}, r)$ in the replay buffer $\mathcal{B}$\;
    
        }

        Sample a transition batch data $\mathcal{D} = \left\{(s,a,s^{\prime},r)\right\}_d$ from replay buffer $\mathcal{B}$\;

        Estimate the $\alpha_{\textrm{CVaR}}$-quantile of the target retrun distribution for each $(s,a)$ in the batch $\mathcal{D}$ by \eqref{eq3-3-2:cvar-term}\;

        Update the policy parameters $\bm{\theta}$ based on $\mathcal{D}$ by \eqref{eq3-18:actor-param-update}\;

        Update the Q-function network parameters $\bm{\phi}$ by minimizing \eqref{eq3-19-1:q-function-loss}\;
        
        Update parameters of target networks $\hat{\bm{\theta}}$ and $\hat{\bm{\phi}}$ by \eqref{eq3-20:soft-update}\;
        
        }
        
	\Return A \textsc{DiffCarl} policy $\pi^*$ with parameters $\bm{\theta}^*$
\end{algorithm}

\section{Results and Discussions}
\label{sec-exp}

In this section, we analyze \textsc{DiffCarl} and demonstrate its outstanding performance by comparing to other existing methods. Besides, we provide insights into the use of \textsc{DiffCarl} and its advantages in MGC energy management.

\subsection{Experimental Setup}
We present the experimental setup in the following aspects.

\subsubsection{Computing Setup}
Our experiments were conducted by a workstation with an NVIDIA GeForce RTX 4090 GPU with 24GB memory and an Intel Core Ultra 7 265KF 20-Core CPU and 64GB of DDR5 RAM. The operating system is Ubuntu 24.04.2 LTS with CUDA V12.2 toolkit. Besides, our experiments are organized with the following details.

\subsubsection{Dataset} 
To evaluate our proposed method's performance, we conducted experiments on multiple datasets, including real-world data and synthetic data. For real-world data we choose the PJM dataset contains publicly available energy system data provided by Pennsylvania-Jersey-Maryland (PJM) Interconnection, one of the largest regional transmission organizations. The PJM dataset contains high temporal resolutions and various market features, and has been widely used by many research studies \cite{huang2020deep, gao2021online}. We adopt following data profiles for our designed MGC environment: 1) solar (PV) generation; 2) wind turbine (WT) generation; 3) load; and 4) real-time energy market locational marginal pricing (LMP). The PJM dataset provides hourly measurements over one year. We first apply preprocessing to eliminate random noise and outlier spikes. To avoid the disturbance of seasonal weather variations, the preprocessed data is sorted monthly, with the first three weeks allocated to the training set and the fourth week reserved for testing. The synthetic dataset is generated based on a nominal data profile, with 20\% white noise added to the real data to introduce variability across samples. Compared to the real-world dataset, the synthetic data spans a narrower value range but exhibits greater randomness in temporal continuity.

\subsubsection{Microgrid Setup} 
We design \textsc{DiffCarl} to be versatile for various MGC settings. Here, we follow the common settings based on existing studies to demonstrate the effectiveness of \textsc{DiffCarl}. The details of the MGC environment parameters can be found in Table \ref{tab1:mgc_param}.

\begin{table}[t]
\centering
\caption{The settings of the MGC environment parameters used in the experimental study.}
\label{tab1:mgc_param}
\renewcommand{\arraystretch}{1.3}
\begin{tabular}{ccc}
\hline\hline
\textbf{Parameters}                                  & \textbf{Description}                                                                         & \textbf{Values}         \\ \hline
$E_{\textrm{ESS}}$                                   & ESS Capacity (kWh)                                                                           & 200                     \\ \hline
$[\eta_{\textrm{ch}},\eta_{\textrm{dis}}]$           & \begin{tabular}[c]{@{}c@{}}ESS Charging and Discharging \\ Coefficiency\end{tabular}         & {[}0.9, 0.95{]}         \\ \hline
$[P_{\textrm{ch}}^{\max},P_{\textrm{dis}}^{\max}]$   & \begin{tabular}[c]{@{}c@{}}ESS Charging and Discharging\\ Power Limit (kW)\end{tabular}      & {[}200, 200{]}          \\ \hline
$[\gamma_{\textrm{ch}}, \gamma_{\textrm{dis}}]$      & \begin{tabular}[c]{@{}c@{}}ESS Charging and Discharging\\ Cost Factor (S\$/kWh)\end{tabular} & {[}0.005, 0.005{]}      \\ \hline
$[E_{\textrm{ESS}}^{\min}, E_{\textrm{ESS}}^{\max}]$ & ESS Storage Limit (kWh)                                                                      & {[}200, 1,800{]}        \\ \hline
$E_{\textrm{ESS}}(0)$                                & ESS Initial Storage (kWh)                                                                    & 1000                    \\ \hline
$[a,b,c]$                                            & CDG Cost Coefficient                                                                         & {[}0.004, 0.066, 0.7{]} \\ \hline
$[P_{\textrm{CDG}}^{\min}, P_{CDG}^{\max}]$          & CDG Power Limit (kW)                                                                         & {[}0, 200{]}            \\ \hline
$R_{\textrm{CDG}}^{\max}$                            & \begin{tabular}[c]{@{}c@{}}CDG Power Maximum \\ Ramp Rate (kW/h)\end{tabular}                & 20                      \\ \hline
$\alpha_{\textrm{LD}}$                               & Maximum Load Shedding Rate                                                                   & 0.5                     \\ \hline
$\lambda_{\textrm{LD}}$                              & Load Shedding Penalty (S\$/kWh)                                                              & 1                       \\ \hline
$\rho_{\textrm{Carbon}}$                             & Carbon Price (S\$/kg CO2)                                                                    & 0.025                   \\ \hline
$[\omega_{\textrm{CDG}}, \omega_{\textrm{Grid}}]$    & Carbon Density (kg CO2/kWh)                                                                  & {[}0.9, 0.412{]}        \\ 
\hline\hline
\end{tabular}
\end{table}

\subsubsection{\textsc{DiffCarl} Model Design} 
\textsc{DiffCarl} follows a soft-actor critics (SAC) framework for MGC energy scheduling with the key novelty of upgrading the actor network with diffusion modelling. We describe the network structure in Table \ref{tab2:model_param} and list the settings of hyper-parameters in Table \ref{tab3:training_param}.

\begin{table}[]
\centering
\caption{The structure of actor and critic networks in \textsc{DiffCarl}.}
\label{tab2:model_param}
\renewcommand{\arraystretch}{1.3}
\begin{tabular}{c|c|c|c}
\hline\hline
\textbf{Networks}       & \textbf{Layers}           & \textbf{Activation} & \textbf{Neurons}        \\ \hline
\multirow{7}{*}{Actor}  & \texttt{SinusoidalPosEmb} & -                   & 16                      \\
                        & \texttt{fc}               & \texttt{mish}       & 32                      \\
                        & \texttt{fc}               & \texttt{mish}       & 16                      \\
                        & \texttt{concat}           & -                   & -                       \\
                        & \texttt{fc}               & \texttt{mish}       & 128                     \\
                        & \texttt{fc}               & \texttt{mish}       & 128                     \\
                        & \texttt{fc}               & \texttt{tanh}       & 45                      \\ \hline
\multirow{3}{*}{Critic} & \texttt{fc}               & \texttt{mish}       & 128                     \\
                        & \texttt{fc}               & \texttt{mish}       & 128                     \\
                        & \texttt{fc}               & -                   & 45                      \\ 
\hline\hline
\end{tabular}
\end{table}

Specifically for the actor and double critic networks, we follow a similar network configuration to mitigate overestimation \cite{haarnoja2018soft}. The structures of the actor and critic networks are described in Table \ref{tab2:model_param}. In the actor network, we use sinusoidal position embedding layer \texttt{SinusoidalPosEmb} to process temporal information in diffusion to derive time embedding vectors. \texttt{concat} allows concatenation of sinusoidal embeddings, denoising step $k$ and state $s$. Besides, the parameters of the networks are optimized by \texttt{adam} optimizer and the training related hyper-parameters are available in Table \ref{tab3:training_param}.

\begin{table}[]
\centering
\caption{The settings of \textsc{DiffCarl} hyper-parameters.}
\label{tab3:training_param}
\renewcommand{\arraystretch}{1.3}
\begin{tabular}{ccc}
\hline\hline
\textbf{Parameters}       & \textbf{Description}                      & \textbf{Values}    \\ 
\hline
$\eta_a$                  & Learning rate of the actor                & $1 \times 10^{-4}$ \\
$\eta_c$                  & Learning rate of the critics              & $1 \times 10^{-3}$ \\
$\tau$                    & Weight of soft update                     & $5 \times 10^{-3}$ \\
$\lambda_{wd}$            & Weight decay                              & $1 \times 10^{-4}$ \\
$\alpha_{\textrm{ent}}$   & Action entropy regularization temperature & $0.05$             \\
$\lambda_{\textrm{risk}}$ & Risk sensitivity coefficient              & $0.1$              \\
$\alpha_{\textrm{CVaR}}$  & Confidence level for computing CVaR       & $0.95$             \\
$\gamma$                  & Discount factor                           & $0.95$             \\
$K$                       & Denoising steps in diffusion model        & $10$               \\
$d$                       & Batch size                                & $256$              \\
$B$                       & Capacity of replay buffer                 & $1 \times 10^{6}$  \\
$E$                       & Maximum training episodes                 & $2000$             \\
$C$                       & Transitions per training step             & $1000$             \\
\hline\hline
\end{tabular}
\end{table}

\subsection{Case Study}
\begin{figure*}[t]
    \centering
    \begin{subfigure}[t]{0.49\linewidth}
        \centering
        \includegraphics[width=\linewidth]{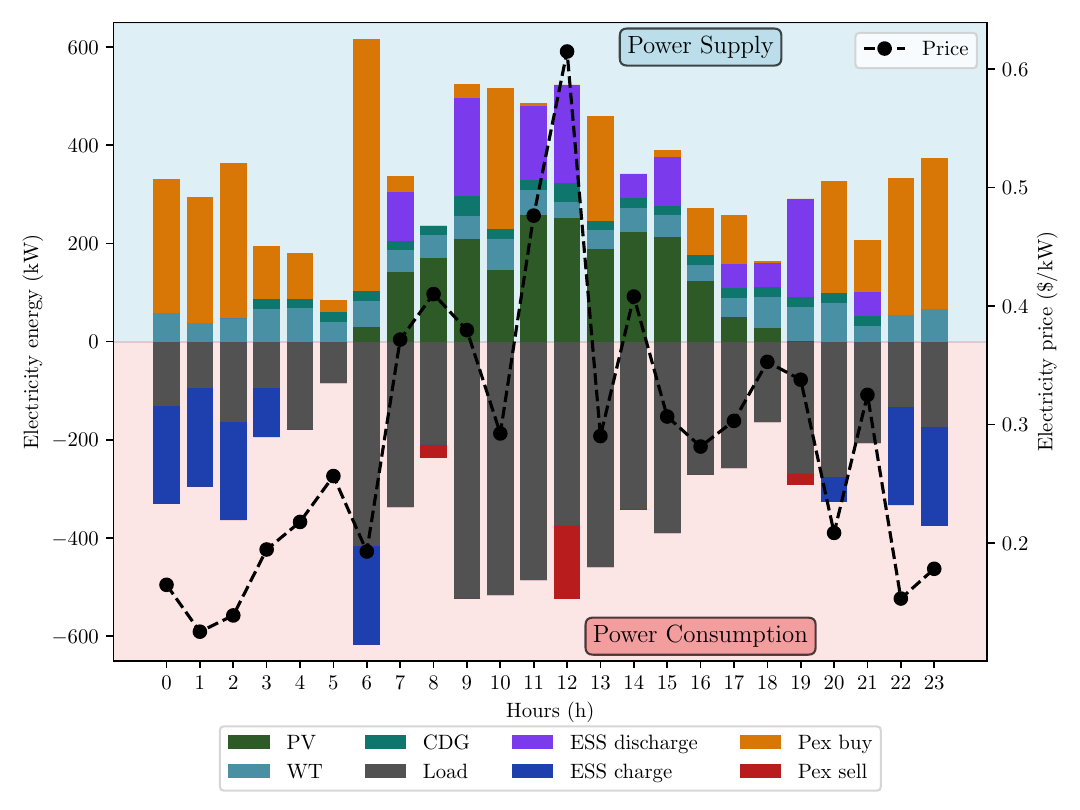}
        \caption{Electrical power balance and electricity price under \textsc{DiffCarl}.}
        \label{fig:case_study_1}
    \end{subfigure}
    \hfill
    \begin{subfigure}[t]{0.49\linewidth}
        \centering
        \includegraphics[width=\linewidth]{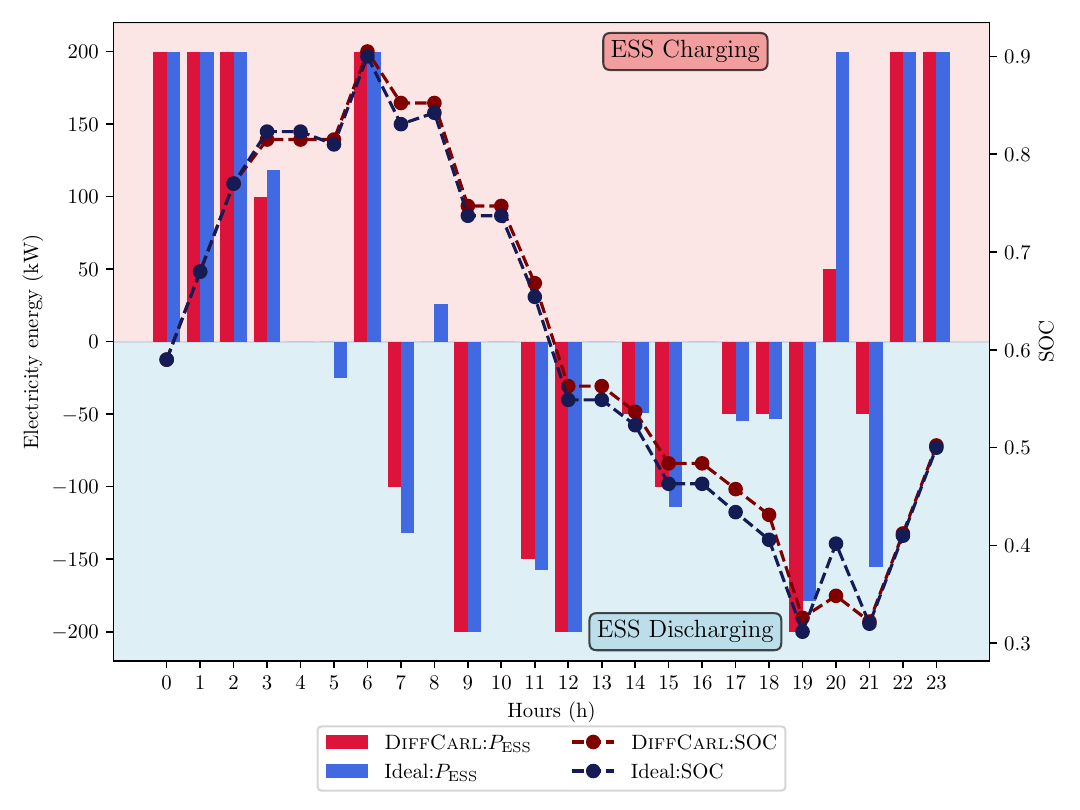}
        \caption{ESS operation and SoC variation under \textsc{DiffCarl} and \textsc{Offline}.}
        \label{fig:case_study_2}
    \end{subfigure}
    \caption{Case study results over a 24-hour test day. (a) shows the electrical power balance including contributions from photovoltaic (PV), wind turbine (WT), community diesel generator (CDG), energy storage system (ESS), and grid interactions (Pex buy/sell), along with the dynamic electricity price under the proposed \textsc{DiffCarl} policy. (b) shows the corresponding ESS operations, including charging and discharging behaviors, and the resulting state-of-charge (SoC) trajectory, compared against an ideal but unrealistic benchmark method \textsc{Offline}.}
    \label{fig:case_study_combined}
\end{figure*}

We first present a case study to illustrate and validate a one-day energy scheduling produced by \textsc{DiffCarl}. The results are shown in Fig. \ref{fig:case_study_combined}, and we have the following observations.

First, \textsc{DiffCarl} can maintain supply-demand balance as shown in Fig. \ref{fig:case_study_1}. In the figure, the positive stacked bars correspond to energy supply, contributed by renewable, CDG, ESS (discharge), and utility grid (purchase). The negative bars represent energy demand, for load, ESS charging, and selling electricity to the grid. In each hour, \textsc{DiffCarl} balances the supply and demand energy while optimizing operational cost and carbon emission. Second, \textsc{DiffCarl} maximizes the utilization of renewable. During periods with high renewable output and low demand, purchases from the utility grid is minimized, or MGC can even sell electricity to the grid. Similarly, \textsc{DiffCarl} tends to purchase more energy when necessary, e.g., with reduced renewable.

The usage of CDG is minimal and adaptive throughout the day. This suggests that CDG is a cost- and carbon-sensitive energy dispatch unit. CDG is activated often when the renewable and ESS are insufficient to well cover the demand. Besides, CDG has lower ramp rate than ESS and the priority of activating it can be lower under urgent energy usage scenarios. ESS serves a similar role as CDG but owns its unique characteristics, e.g., large maximum ramp rate and non-carbon-intensive. Seen from Figure \ref{fig:case_study_combined}, ESS is being charged during off-peak hours with low electricity prices. Similarly, discharge happens more often during peak hours with high prices. Overall, this case study demonstrates clearly that \textsc{DiffCarl} is able to produce a valid and effective schedule in operating the MGC system. 

\textsc{DiffCarl} deals with uncertain future system dynamics. We would like to further compare its performance with the best possible scheduling. Here, we introduce \textsc{Offline}, which assumes full knowledge of future events and performs offline optimal decision-making to provide the theoretical upper bound of the optimization performance. Note that \textsc{Offline}'s assumption is unrealistic as the future dynamics of the system is unknown or uncertain and the decision-making cannot be in an offline manner. In Fig. \ref{fig:case_study_2}, ESS output and its SoC profiles are compared between \textsc{DiffCarl} and \textsc{Offline}. Both strategies exhibit a clear pattern of charging at night and early hours with low prices or demand, and discharging during daytime and peak hours. \textsc{DiffCarl} achieves very similar charging/discharging profiles compared to \textsc{Offline}. Besdies, we notice that \textsc{DiffCarl} is not aggressive in ESS operations, for its awareness of risks and potential penalties on violating the SoC constraints.

\subsection{Comparison Study} 
We compare \textsc{DiffCarl} with several comparison algorithms to demonstrate its effectiveness in energy scheduling.

\subsubsection{Comparison Algorithms} 
We consider a few benchmark algorithms as the baselines, including DQN, DDPG and SAC where technical details can be found in literature. We also implement several methods including \textsc{Offline} described above and the others detailed below.

\textsc{Day-ahead}:
This algorithm does not make unrealistic assumptions as \textsc{Offline}. It makes predictions of the next-day system dynamics and aims to make the best decisions based on the predictions, which, are not the ground-truth.

\textsc{Myopic}:
This method is based on greedy algorithm and decisions are made step-by-step. In each step, the method collects available information of the step and optimizes the decision-making of the step without considering future consequences.
    
\textsc{MPC}: This method addresses the limitations of the above algorithms by incorporating both future predictions and real-time feedback into its decision-making. Predictions are assumed to be more accurate for near-future and less accurate for long-future.

\subsubsection{Operating Cost}
We define a relative cost metric to compare the performance of a given algorithm against \textsc{DiffCarl} as
\begin{equation}
\label{eq4-1:rel_cost}
    \frac{ C_{\textsc{DiffCarl}} - C_{\textsc{comp}}}{C_{\textsc{DiffCarl}}} \times 100\%,
\end{equation}
where $C_{\textsc{comp}}$ and $C_{\textsc{DiffCarl}}$ are the costs of the comparison algorithm and \textsc{DiffCarl}, respectively. The improvement is represented as a percentage and a negative value means that \textsc{DiffCarl} is better than the comparison algorithm, and vice versa. The comparison results are shown in Table \ref{tab3:performance_comparison}. 

\begin{table*}[]
\centering
\caption{Performance comparisons between \textsc{DiffCarl} and different benchmarks. Negative percentages mean the comparison algorithms are outperformed by \textsc{DiffCarl}. \textsc{DiffCarl} can well approximate the optimal performance by \textsc{Offline}. Compared to the rest of other algorithms, it achieves the best performance for all scenarios including 2MG, IEEE 15-bus and 33-bus in terms of both overall cost and carbon emission.}
\label{tab3:performance_comparison}
\renewcommand{\arraystretch}{1.3}
\begin{tabular}{c|ccr|ccr|ccr}
\hline\hline
\multirow{2}{*}{Algorithms} & \multicolumn{3}{c|}{2MG Case}      & \multicolumn{3}{c|}{IEEE 15-bus}   & \multicolumn{3}{c}{IEEE 33-bus}    \\ 
\cline{2-10}             & Cost   & Carbon &  \multicolumn{1}{c|}{Improv.}     & Cost        & Carbon  & \multicolumn{1}{c|}{Improv.}    & Cost         & Carbon & \multicolumn{1}{c}{Improv.}       \\ \hline
\textsc{Day-ahead}       & 965.30 & 92.24  &  $-$30.12\%  & 3,273.77    & 198.19  & $-$26.67\% & 9,267.06     & 686.81 &  $-$26.79\%   \\ 
\textsc{Myopic}          & 872.35 & 88.07  &  $-$17.59\%  & 3,084.49    & 197.31  & $-$19.35\% & 8,916.11     & 642.75 &  $-$21.99\%   \\ 
\textsc{MPC-8}           & 801.36 & 67.77  &  $-$8.02\%   & 2,891.45    & 171.07  & $-$11.88\% & 8,521.62     & 581.18 &  $-$16.59\%   \\ 
DQN                      & 789.21 & 64.72  &  $-$6.38\%   & 2,835.14    & 160.62  & $-$9.70\%  & 8,351.87     & 553.97 &  $-$14.27\%   \\ 
SAC                      & 762.14 & 63.52  &  $-$2.73\%   & 2,701.24    & 154.41  & $-$4.52\%  & 8,065.71     & 532.94 &  $-$10.36\%   \\ 
DDPG                     & 758.80 & 61.21  &  $-$2.28\%   & 2,609.01    & 148.26  & $-$0.95\%  & 7,551.94     & 498.84 &  $-$3.32\%    \\ 
\hline
\textsc{DiffCarl} (ours) & \textbf{741.86}    & \textbf{59.25}   & \multicolumn{1}{c|}{$-$} & \textbf{2,584.46}    & \textbf{144.75}  & \multicolumn{1}{c|}{$-$} & \textbf{7,308.61}    & \textbf{483.95}    & \multicolumn{1}{c}{$-$} \\ \hline
\textsc{Offline}         & 724.84 & 54.84  & 2.29\%       & 2,461.49    & 131.55  & 4.76\%     & 7,064.51     & 457.38 & 3.34\%        \\ 
\hline\hline
\end{tabular}
\end{table*}

First, let us consider an MGC scenario with 2 microgrids, represented as 2MG in the table. It consists of 2 loads, 2 PVs, 2 WTs, 2 CDGs and 2 ESS units. Compared to all the tested online algorithms, \textsc{DiffCarl} achieves the lowest cost and carbon emission. Overall, \textsc{DiffCarl}'s cost performance is significantly better than other baselines, and \textsc{DiffCarl} outperforms \textsc{Day-ahead} the most with $30.1\%$ improvement. We further compare \textsc{DiffCarl} with the unrealistic \textsc{Offline} with perfect and optimal energy scheduling. The results show that \textsc{DiffCarl} can approximate the optimal performance by $98.2\%$, which aligns with our observation in the case study that \textsc{DiffCarl} develops a near-optimal policy. A reason for \textsc{DiffCarl}'s competitive performance is that the diffusion model in the actor network is effective in approximating the potentially complex distribution of the optimal actions.

Among the comparison algorithms, \textsc{Day-ahead}, \textsc{Myopic}, and \textsc{MPC} are not learning-based and their performance is relatively not competitive. \textsc{Day-ahead} is basically a static algorithm and performs poorly due to its lack of adaptability. \textsc{Myopic} and \textsc{MPC} are considered as online methods which perform scheduling based on real-time information. Their performance is constrained by the prediction accuracy and scheduling horizon. As a result, they are only capable to deal with scenarios that are similar to the nominal data profile, e.g., with less challenging and more accurate predictions, and the performance with real-world dataset suffers. \textsc{Myopic} and \textsc{MPC} both lack temporal generalization or adaptability to uncertainty and variability compared to diffusion-based \textsc{DiffCarl}, and the performance gaps are $17.6\%$ and $8.0\%$, respectively.

RL-based algorithms have been the state-of-the-art for energy scheduling, and they outperform the non-RL algorithms presented above. Our proposed \textsc{DiffCarl} manages to produce improved scheduling compared to these competitive algorithms, including DQN, SAC and DDPG. Among these algorithms, DQN performs the worst with a $6.4\%$ gap. DQN is a conventional method and it emulates table-seeking Q-learning algorithm which is poorly suited for fine-grained problems like MGC energy scheduling. SAC and DDPG perform better, but still worse than \textsc{DiffCarl} with $2.7\%$ and $2.3\%$ higher costs, respectively. RL-based methods are more adaptive to unprecedented scenarios and data variance due to its learning based pattern by actively interacting with the MGC environment. \textsc{DiffCarl}, as an enhanced RL, not only inherits the RL advantages, but also captures temporal awareness with its generative mechanism based on diffusion modelling.

We further investigate the algorithms' performance under two standard IEEE test feeders, including IEEE 15-bus and 33-bus systems. The former has 10 loads, 2 PVs, 3 ESS units, and 2 CDGs, and the latter consists of  26 loads, 5 PVs, 7 ESS units, and 5 CDGs. The results are summarized in Table \ref{tab3:performance_comparison}. The results demonstrate that \textsc{DiffCarl}'s competitiveness is not limited to the 2MG case and the algorithm can outperform comparison algorithms consistently across different power system scales, e.g., under IEEE 15-bus and 33-bus systems. Compared to traditional scheduling strategies including \textsc{Day-ahead}, \textsc{Myopic}, and \textsc{MPC}, \textsc{DiffCarl} achieves substantial reductions in both operational cost and carbon emissions, with cost savings exceeding $15\%$. While RL-based methods such as DQN, SAC and DDPG show improved efficiency compared to numerical methods, \textsc{DiffCarl} still surpasses them in IEEE 33-bus system by $14.3\%$, $10.4\%$ and $3.3\%$, respectively.

In summary, the results show that \textsc{DiffCarl} remains robust as system complexity increases, indicating strong scalability and generalization capability. The results validate \textsc{DiffCarl} as a highly effective and sustainable framework for intelligent energy management in MGC systems.

\subsubsection{Learning and Convergence}
\begin{figure}[!t]
    \centering
    \includegraphics[width=\linewidth]{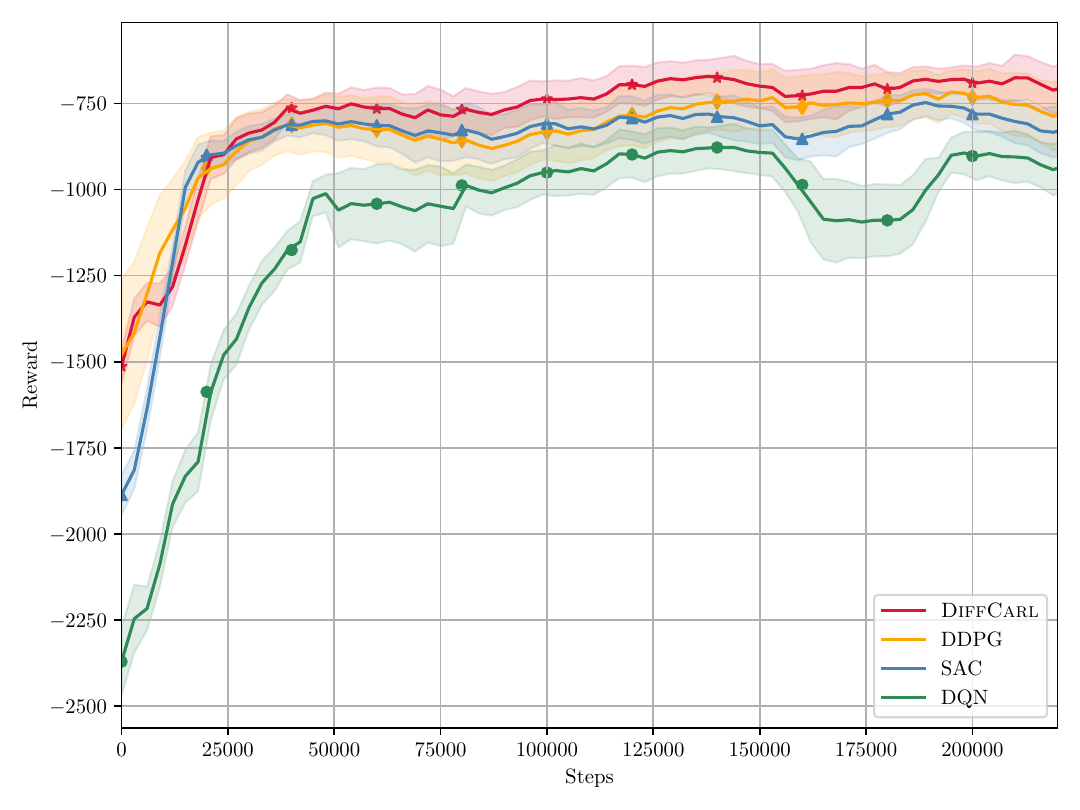}
    \caption{A comparison of learning curves under different RL policies. \textsc{DiffCarl} converges fast and stably outperforms other RL policies with higher test reward.}
    \label{fig:learning_curve}
\end{figure}

\textsc{DiffCarl}, together with DQN, SAC, and DDPG, are RL-based and learn optimal policies by maximizing the reward during training. We visualize learning curves of these algorithms in Fig. \ref{fig:learning_curve} with 250k steps. Seen from the figure, \textsc{DiffCarl} converges to the largest reward at the end of training. During training process, \textsc{DiffCarl}'s best test reward is $-666.5$ which is higher than DQN, SAC and DDPG. Among all benchmarks, \textsc{DiffCarl} brings the best performance overall, consistently yielding the highest test reward throughout training. Meanwhile, its fast convergence allow the agent to swiftly get a well-trained policy even before 50k steps. Lastly, the smooth curve with minimal fluctuations indicates a very robust and stable learning process of \textsc{DiffCarl}.

\textsc{DiffCarl} is able to achieve higher rewards during most of the training process, indicating that its policy outperforms the policies of the other RL-based methods. One of the most inspiring features of the diffusion modeled RL is that a noise predictor is applied to generate actions, which significantly helps to develop a satisfactory policy even in unprecedented scenarios, eliminating the randomness in action generation and reducing high-cost operations in scheduling.

\subsection{\textsc{DiffCarl} Carbon- and Risk-awareness}
\textsc{DiffCarl} not only minimizes the total cost with optimized energy scheduling, but also achieves carbon- and risk-awareness. In this part, we specifically present the results of \textsc{DiffCarl}'s carbon emission and risk-aware scheduling.

\subsubsection{Carbon Emission}
\begin{figure}[!t]
    \centering
    \includegraphics[width=\linewidth]{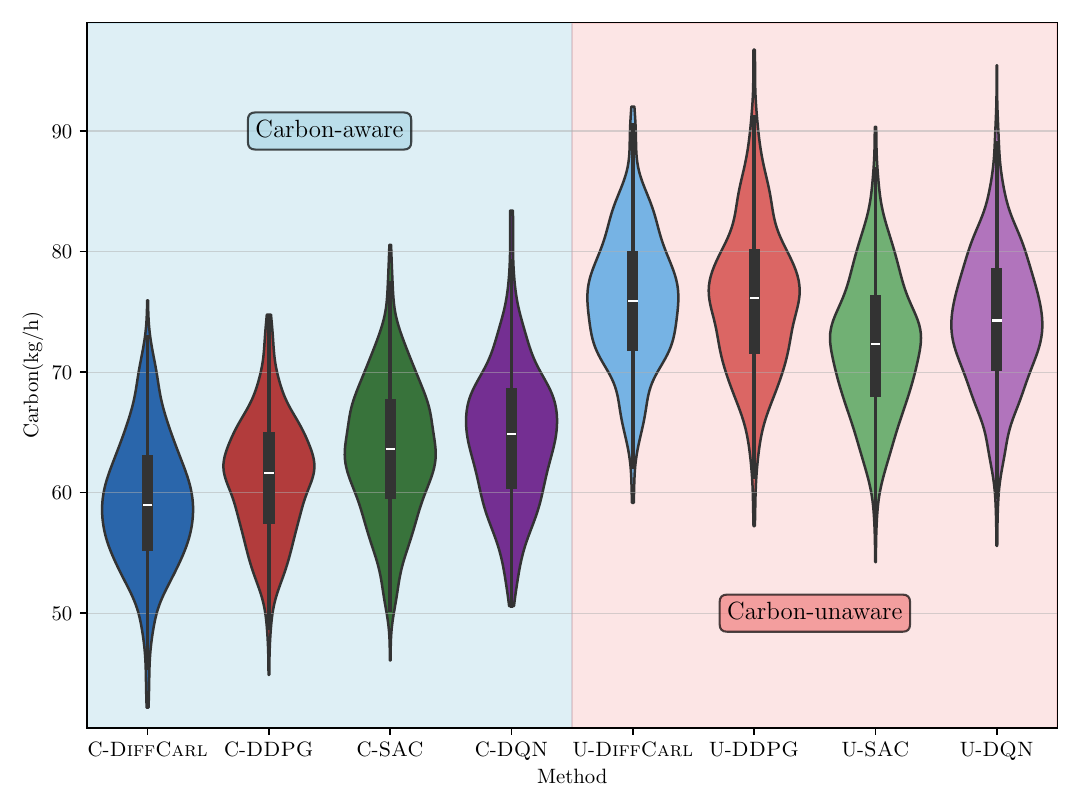}
    \caption{Carbon emission distributions under carbon-aware (C-) and carbon-unaware (U-) policies. Carbon-aware (C-) methods show lower carbon emission with less variance than their carbon-unaware (U-) alternatives. \textsc{DiffCarl} shows the lowest carbon emission among carbon-aware (C-) policies.}
    \label{fig:carbon_distribution_comparison}
\end{figure}

Carbon emission is part of the formulated problem in this study. As such, the comparison algorithms incorporating the same cost functions are essentially carbon-aware, same as \textsc{DiffCarl}. Here, we present \textsc{DiffCarl}'s carbon emission compared to the rest RL-based algorithms that are carbon-aware. We further demonstrate the impact of removing carbon factors from the problem formulation and present the results of non-carbon-aware algorithms. Fig. \ref{fig:carbon_distribution_comparison} shows the carbon emission of different tested algorithms and some numerical results are available in Table \ref{tab3:performance_comparison}.

First, carbon shall be reflected in energy scheduling problems. Among all RL policies, the carbon-aware group methods have consistently lower carbon emissions than their carbon-unaware counterparts. Carbon-unaware methods show wider and higher distribution tails, especially U-DDPG and U-DQN, suggesting greater variability and higher average emissions. Compared with U-\textsc{DiffCarl}'s carbon emission $64.7$ kg/h, C-\textsc{DiffCarl} has much less emission with $59.3$ kg/h, which indicates approximately $28.7\%$ less carbon, significantly promoting a greener operation of microgrids. Meanwhile, it is obvious that all listed carbon-aware methods allow the MGC to produce much less carbon than their carbon-unaware alternatives. This indicates how algorithms' carbon-awareness reduces environmental impact during MGC operations and thereby contributing to sustainable development.

Among the carbon-aware RL algorithms, \textsc{C-DiffCarl} outperforms the rest regarding carbon emission. Our algorithm has the lowest median carbon emission $58.9$ kg/h with a narrow distribution, demonstrating its reliability in minimizing carbon emission. In conclusion, the results highlight the importance of carbon-awareness for minimizing carbon emission overtime, and the carbon-aware \textsc{DiffCarl} is the most robust and effective among the tested algorithms.

\subsubsection{Risk Sensitivity}
\begin{figure}[!t]
    \centering
    \includegraphics[width=\linewidth]{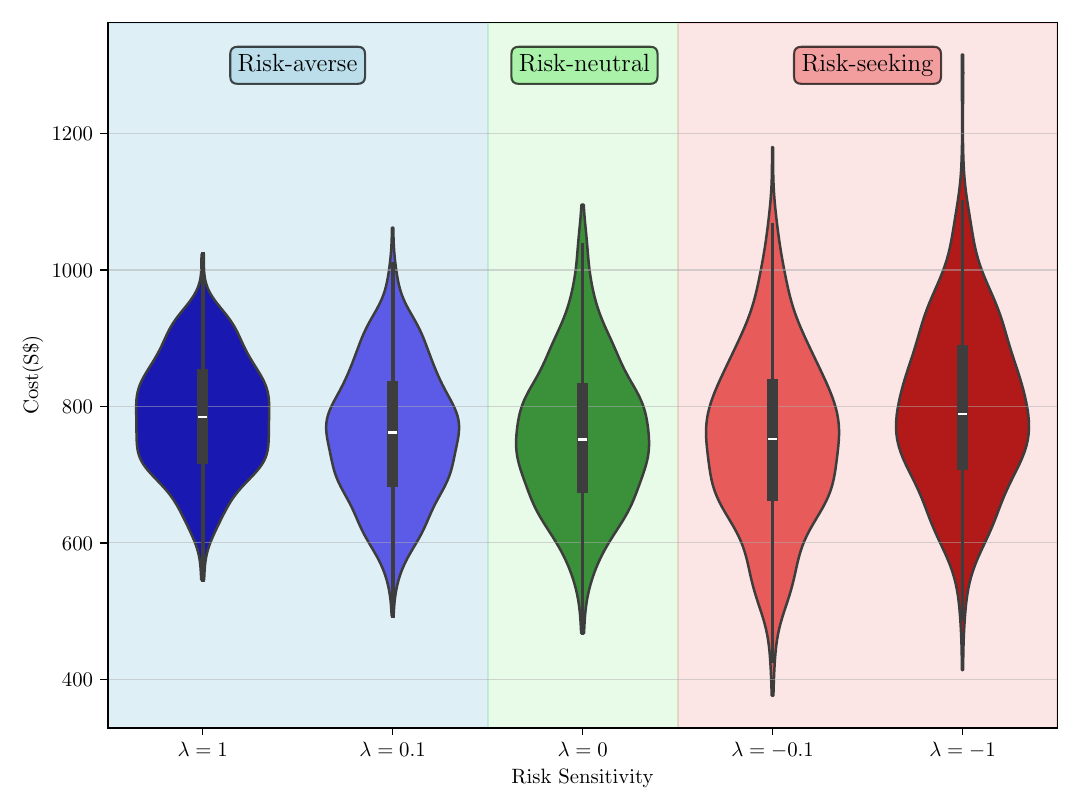}
    \caption{Cost distributions under \textsc{DiffCarl} policies with varying risk-sensitivity parameters. $\lambda>0$ indicates risk-averse behavior with tighter cost distributions, while $\lambda<0$ indicates risk-seeking behavior with broader cost distributions.}
    \label{fig:risk_distribution_comparison}
\end{figure}

\textsc{DiffCarl} is risk-aware and we discuss the relationship between cost and risk in this part. We applied CVaR to eliminate worst case outcomes with a high risk of introducing high operational costs. \textsc{DiffCarl}'s sensitivity on risks can be controlled by a risk parameter $\lambda$ based on Eq. \eqref{eq3-3-1:rl-risk}, where a positive $\lambda$ indicates a risk-averse strategy while a negative value implies risk-seeking. Here, we consider three levels of risk sensitivities, including risk-averse, risk-neutral, and risk-seeking \cite{eriksson2019epistemic}. The results are shown in Fig. \ref{fig:risk_distribution_comparison} and the discussions are as below.

The results reveal clear trade-offs between cost optimization and risk management in \textsc{DiffCarl} policies. Seen from the figure, risk-seeking leads to the most significant cost fluctuations, e.g., with a long distribution. This implies that the pursuit of high-risk cost minimization comes with the increase of system instability and variability. Meanwhile, risk-averse policies provide more predictable costs at a moderate premium. It suggests that moderate risk-seeking behavior does not influence cost efficiency much, but significantly increase the system stability.

When examining median costs across different risk sensitivity levels, the risk-neutral emerges as the top performer with cost S\$751.8, followed by risk-seeking policy with $\lambda=-0.1$ with nearly the same cost, S\$752.0. Risk-averse policies command slightly higher costs. When $\lambda = 0.1$, the cost is S\$761.7, and the gap between risk-neutral and risk-averse is about S\$10. When $\lambda = 1$, the gap increases significantly to S\$33. A very aggressive risk-seeking policy with $\lambda = -1$ performs poorly with the highest median cost S\$789.0, suggesting that extreme risk-seeking behavior is counterproductive and instable and \textsc{DiffCarl}'s risk-awareness is necessary.

Risk-averse policies demonstrate their primary value proposition through significantly reduced volatility. The risk-averse policy of $\lambda = 1$ achieves the most stable performance with a standard deviation of only S\$84.1, representing a 21\% reduction compared to the neutral policy's S\$106.4. While a moderate policy with $\lambda = 0.1$ provides moderate stability improvement with a standard deviation of S\$97.0, offering a 9\% volatility reduction. These policies also provide better worst-case protection, with maximum costs ranging from S\$1,024 to S\$1,062 compared to S\$1,096 for risk-neutral. Risk-seeking policies present a more uncertain and instable feature. Both risk-seeking variants show approximately 18\% higher volatility than risk-neutral, and their worst case costs range from S\$1,180 to S\$1,316, which brings up extra concern in real-world energy applications.

Overall, experiments demonstrate that risk-averse policy with $\lambda=0.1$ emerges as the optimal policy for most applications, successfully balancing cost efficiency with manageable risk exposure. For risk-constrained environments, a more conservative policy of $\lambda=1$ provides excellent volatility reduction at a reasonable cost premium, making it ideal for applications requiring predictable and acceptable budgets. The results reveal that moderate risk preferences tend to outperform extreme positions by striking a balance in eliminating uncertainty and optimizing cost, which maximizes the effectiveness of \textsc{DiffCARL} policies. In reality, the users may choose and adopt their own risk sensitive policy according to their needs, with a customized emphasis on safety or performance.

\section{Conclusion}
\label{sec-conclusion}
In this paper, we have proposed \textsc{DiffCarl}, a diffusion-modeled carbon- and risk-aware reinforcement learning algorithm for microgrid energy scheduling and optimization. \textsc{DiffCarl} enables grid operators to schedule microgrids effectively while promoting sustainability and robustness. Integrating a diffusion-based policy framework with explicit carbon- and risk-sensitive objectives balances operational efficiency with environmental and reliability constraints. Extensive experiments have demonstrated that \textsc{DiffCarl} achieves 2.3$–$30.1\% lower operational costs than baseline methods, reduces carbon emissions by 28.7\% compared to its carbon-unaware variant, and shows lower performance variability under dynamic and uncertain conditions. These results have underscored the potential of GenAI-based models in advancing intelligent and sustainable energy systems. In the future, we would like to investigate the applicability of \textsc{DiffCarl} to different microgrid configurations and settings. We also plan to explore \textsc{DiffCarl}'s adaptation to related domains such as battery energy storage management and smart manufacturing, where uncertainty, sustainability, and dynamic scheduling are also critical.
\balance

\bibliographystyle{IEEEtran}
\bibliography{DiffCarl}


 




\vfill

\end{document}